%% file: main.tex
\pdfoutput=1
\UseRawInputEncoding
\documentclass[conference]{IEEEtran}
\IEEEoverridecommandlockouts
\usepackage{cite}
\usepackage{amsmath,amssymb,amsfonts}
\usepackage{algorithmic}
\usepackage{graphicx}
\usepackage{textcomp}
\usepackage{caption}
\usepackage{multirow}
\usepackage{bm}
\usepackage{booktabs}
\usepackage{bbold}

\def\BibTeX{{\rm B\kern-.05em{\sc i\kern-.025em b}\kern-.08em
    T\kern-.1667em\lower.7ex\hbox{E}\kern-.125emX}}
\makeatletter
\let\NAT@parse\undefined
\makeatother
\usepackage{hyperref}

\begin{document}

\title{OSIS: Efficient One-stage Network for 3D Instance Segmentation
}

\author{\IEEEauthorblockN{1\textsuperscript{st} Chuan Tang}
\IEEEauthorblockA{\textit{School of Artificial Intelligence} \\
\textit{Jilin University}\\
Changchun, China \\
tangchuan20@mails.jlu.edu.cn}

\and

\IEEEauthorblockN{2\textsuperscript{nd} Xi Yang}
\IEEEauthorblockA{\textit{School of Artificial Intelligence} \\
\textit{Jilin University}\\
Changchun, China \\
yangxi21@jlu.edu.cn}
}

\maketitle

\begin{abstract}
Current 3D instance segmentation models generally use multi-stage methods to extract instance objects, including clustering, feature extraction, and post-processing processes. 
However, these multi-stage approaches rely on hyperparameter settings and hand-crafted processes, which restrict the inference speed of the model. 
In this paper, we propose a new 3D point cloud instance segmentation network, named OSIS. 
OSIS is a one-stage network, which directly segments instances from 3D point cloud data using neural network. 
To segment instances directly from the network, we propose an instance decoder, which decodes instance features from the network into instance segments. Our proposed OSIS realizes the end-to-end training by bipartite matching, 
therefore, our network does not require computationally expensive post-processing steps such as non maximum suppression (NMS) and clustering during inference.
The results show that our network finally achieves excellent performance in the commonly used indoor scene instance segmentation dataset, and the inference speed of our network is only an average of 138ms per scene, which substantially exceeds the previous fastest method.
\end{abstract}

\begin{IEEEkeywords}
instance segmentation, 3D, point cloud
\end{IEEEkeywords}

\input{sections/1_intro}
\input{sections/2_related}
\input{sections/3_method}
\input{sections/4_exp}

\input{sections/5_conclusion}

% \input{ref.tex}
% \bibliographystyle{ieeetr}
% \bibliography{main}

\bibliographystyle{IEEEtran}
\bibliography{IEEEexample}

\end{document}

%% file: sections/1_intro.tex
\section{Introduction}

With the widespread popularity of 3D sensors such as Lidar, Radar, and RGB-D Camera in some fields like VR/AR, autonomous driving, robot, and HD (High Definition) maps, more and more researches focus on 3D scene understanding, including 3D object detection~\cite{Wang2021FCOS3DFC, Zhang2020H3DNet3O}, 3D segmentation~\cite{Han2020OccuSegO3,Chen2021HierarchicalAF,He2021DyCo3DRI,Vu2022SoftGroupF3,Jiang2020PointGroupDP,Engelmann20203DMPAMA,Wang2019AssociativelySI}, 3D object tracking~\cite{Pang2021SimpleTrackUA}, and 3D multimodal understanding~\cite{text2shape, Tang2021Part2WordLJ}, etc. 
Many 3D point cloud instance segmentation works have appeared in recent years, and have achieved extraordinary results. 
3D point cloud instance segmentation aims to extract instance-level objects from the 3D environment and obtain the semantic categories of the instance. 

The current 3D instance segmentation methods are mainly divided into two paradigms, one is proposal-based segmentation, which first uses the existing 3D object detector to extract the object regions, and then applies the mask prediction to obtain the instance masks and instance semantic categories from the local object regions. 
This paradigm relies on heavy region proposal processing such as 3D detector, the inference process of the entire model can hardly improve the efficiency. 
Another paradigm is cluster-based segmentation, which treats the 3D instance segmentation task as clustering from point cloud data, which directly extracts instances from points. 
Meanwhile, to obtain the semantic category of the instance and improve the accuracy of the instance prediction, 
such a paradigm often applies a post-processing network or a hierarchical clustering algorithm for further scoring and refinement. 
However, this paradigm relies on a heuristic grouping algorithm, which is very sensitive to hyper-parameters such as clustering radius, and it is challenging to balance between over-segmentation and under-segmentation. Moreover, clustering-based methods often cause over-segmentation of instances for some instances composed of several non-connected parts. 
Although the previous methods have achieved excellent performance, these multi-stage methods make these methods unable to meet the requirements of real-time scenarios.

\begin{figure}[t]
\centering
\includegraphics[width=1.0\linewidth]{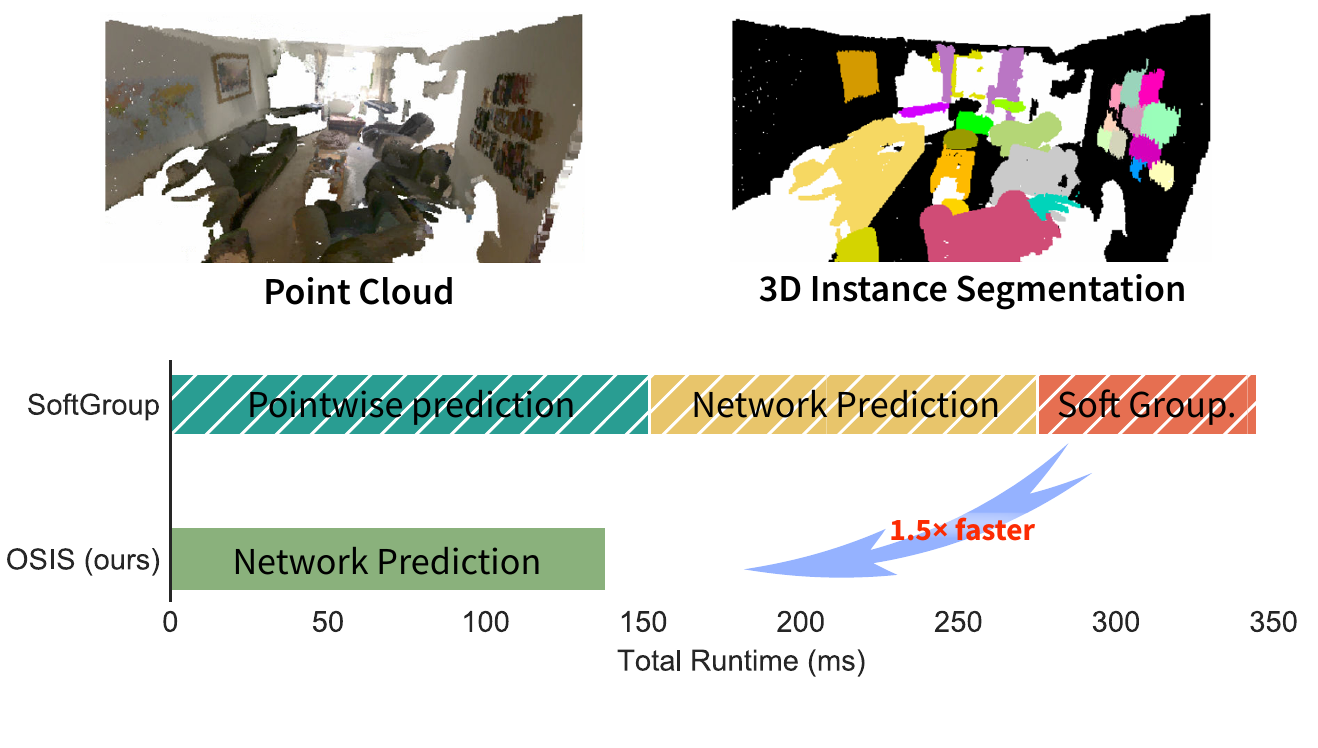}
\caption{We propose a one-stage 3D instance segmentation network, which directly segments instances from 3D point cloud data using neural network. Inference speed increases by 150\% compared with the SOTA method.}
\label{fig:teaser}
\end{figure}

In this paper\footnote{Also see \cite{Liu20223DQueryISAQ,Schult2022Mask3DF3,PointInst}
for concurrent work that proposes similar ideas.}, we propose a one-stage 3D instance segmentation network called OSIS, which predicts instance masks directly from the network. 
We focus on addressing two major challenges. The first is that the complexity of directly extracting instance masks from the entire point cloud is more complex, and it is more difficult to achieve accurate results.
The second is that compared with the proposal-based method which often extracts only one instance object in a region proposal, the one-stage network is more susceptible to the interference of surrounding instances when extracting instance masks.
To address these challenges, we use the instance decoder module to predict instance masks in the 3D point cloud scene.
Our network first generates a set of masks dynamically from the backbone as initial instance masks.
Then, we use the proposed mask-to-feature module to extract instance features.
Finally, the instance features are used as convolution kernels to perform dynamic convolution with the voxelized feature map, generating a new instance mask and feature.
In addition, our network does not use non-maximum suppression (NMS) to filter out redundant instances. 
Instead, we remove duplicate candidate instances during training using bipartite matching to establish a one-to-one mapping relationship between candidate instances and ground truth annotations.
The main contributions of this paper are as follows:

\begin{itemize}
    \item We propose a new one-stage 3D point cloud instance segmentation network, which does not require computationally expensive pre-processing and post-processing, and obtains instance segmentation prediction results directly from the network output, which greatly improves the network inference speed.
    \item In our network, the instance decoder decodes instances from the point-wise features, which gets rid of the dependence on hand-crafted algorithms.
    \item The experimental results show that our network can achieve good instance segmentation accuracy while significantly shortening the network inference time compared to previous work, providing a baseline framework for real-time 3D instance segmentation.
\end{itemize}

%% file: sections/2_related.tex
\section{Related Work}

\subsection{3D Instance Segmentation}
The current mainstream paradigm is mainly divided into two paradigms: proposal-based method~\cite{Hou20193DSIS3S, Yi2019GSPNGS, Yang2019LearningOB} and cluster-based method~\cite{Wang2019AssociativelySI, Jiang2020PointGroupDP, He2021DyCo3DRI, Vu2022SoftGroupF3, Han2020OccuSegO3, Wu2020SSTNetDM, Chen2021HierarchicalAF}. The proposal-based segmentation methods use the existing 3D object detector to extract the object regions, and then apply the mask prediction to obtain the instance masks and instance semantic categories from the local object regions. The cluster-based segmentation methods gather points to obtain instances, these methods often adopt hybrid and hierarchical clustering strategies to achieve higher segmentation quality.
Among these methods, Jiang et al.\cite{Jiang2020PointGroupDP} proposed PointGroup to cluster instances from points by predicted instance centroids and semantic categories. He et al.~\cite{He2021DyCo3DRI} proposed DyCo3D to extract features from the proposals obtained by clustering as convolution kernels and obtains instance masks through dynamic convolution. HAIS~\cite{Chen2021HierarchicalAF} introduces hierarchical aggregation to progressively generate instance proposals. SoftGroup~\cite{Vu2022SoftGroupF3} pays attention to the obstacles of semantic segmentation errors to the performance of instance segmentation, and adopts an adaptive threshold strategy to eliminate the obstacles from semantic segmentation prediction errors. Moreover, OccuSeg~\cite{Han2020OccuSegO3} and SSTNet~\cite{Wu2020SSTNetDM} reduce the complexity of clustering by preprocessing input points into super-voxel and greatly improve the accuracy of the model, however, the super-voxel preprocessing method itself is extremely time-consuming.

\subsection{Dynamic Convolution}

Different from the vanilla convolution operation, the dynamic convolution can conditionally generate feature maps~\cite{Chen2019DynamicCA, Yang2019CondConvCP}. 
SOLO v2~\cite{wang2020solov2} designed a simple, straightforward and fast 2D instance segmentation model, which decouples the mask prediction into the dynamic kernel and convolutional features. 
CondInst~\cite{Tian2020ConditionalCF} uses dynamic convolution for object mask generation. 
SparseInst~\cite{Cheng2022SparseIA} generates instance features from activation maps for instance segmentation and applies dynamic convolution to generate instance masks. 
SOLQ~\cite{Dong2021SOLQSO} explores an end-to-end instance segmentation solution by learning a unified query representation without any post-processing procedures. 
It is worth noting that dynamic convolution is also widely used in the unified segmentation framework (semantic segmentation, instance segmentation, and panoptic segmentation)~\cite{Cheng2021MaskFormer, Cheng2022Mask2Former, Zhang2021KNetTU}. 
For the 3D instance segmentation task, DyCo3D~\cite{He2021DyCo3DRI} first proposed to realize mask generation through dynamic convolution. However, this method does not get rid of the dependence on the clustering process and cannot achieve one-stage inference.

\subsection{Bipartite Matching}

Label assignment is a key problem for training object detection and segmentation models, 
most previous methods employ greedy algorithms to match predicted instances with ground truth,
 however, this approach often makes redundant predictions during inference.
 Therefore, it is very important to provide a suitable target for each proposal during training. Recently, some methods\cite{Erhan2013ScalableOD, Carion2020DETR, wang2020solov2, Cheng2021MaskFormer, Cheng2022Mask2Former, Zhang2021KNetTU} have tried to apply bipartite matching to computer vision tasks such as 2D object detection and segmentation, and have achieved distinguished results. In 3D point cloud instance segmentation, 3D-BoNet~\cite{Yang2019LearningOB} also uses Hungarian matching training to associate proposal bounding boxes with targets.

 In our network, we use bipartite matching to assign the mapping relation between ground truth and proposal instances during training. Therefore, our method could produce few redundant instance predictions even if removing NMS post-processing.

%% file: sections/3_method.tex
\section{Our Method}

\begin{figure*}[t]
    \begin{center}
    \includegraphics[width=1.0\linewidth]{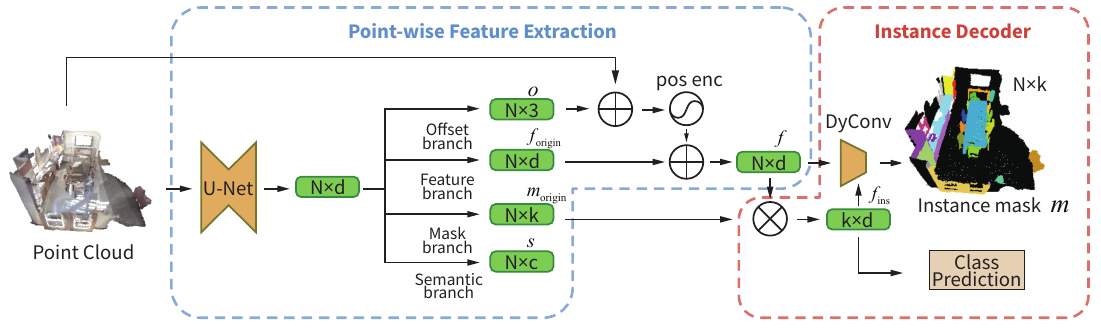}
    \end{center}
    \caption{Illustration of our proposed OSIS comprises a point-wise feature extraction module and an instance decoder. The point-wise feature extraction module is designed to extract point-wise features from the point cloud data, while the instance decoder generates instance-level features and masks using dynamic convolution. }
    \label{fig:overview}
\end{figure*}

Figure~\ref{fig:overview} provides an overview of our proposed network, consisting of two modules: a point-wise feature extraction module and an instance decoder.
We use the sparse convolution framework~\cite{sparseconv} to build the point-wise feature extraction module, and the instance decoder is designed to predict instance segmentation results. Additionally, we introduce the training loss of our network, which involves establishing a one-to-one mapping relationship between candidate instances and ground truth annotations during training using bipartite matching and combining the multi-task loss to train the entire network.
Finally, we provide a description of the inference process used by our network.

\subsection{Point-wise Feature Extraction}

For the input point cloud $P\in\mathbb{R}^{N\times (3+C)}$, where $N$ represent the point number, $C$ denotes the dimension of input point feature, 
First, we convert the point cloud data into volumetric grids $V\in\mathbb{R}^{N^{'}\times (3 + C)}$, where $N^{'}$ denotes the number of voxels. Then we feed it into the backbone of U-Net~\cite{ronneberger2015unet} network with sparse convolution and sparse manifold convolution to extract point-wise features. The output features obtained through the backbone will be sent into four parallel branches: offset branch, feature branch, mask branch and semantic branch.
Through these four branches, we can obtain 
centroid offset $\bm{O}=\{\bm{o}^i\}_{i=1}^N\in\mathbb{R}^{N\times 3}$, 
original point-wise features $\bm{F}=\{\bm{f}_{\text{origin}}^i\}_{i=1}^N\in\mathbb{R}^{N\times d}$,
original instance mask features $\bm{M}=\{\bm{m}_{\text{origin}}^j\}_{j=1}^k\in\mathbb{R}^{N\times k}$, 
and point-wise semantic predictions $\bm{S}=\{\bm{s}^i\}_{i=1}^N\in\mathbb{R}^{N\times c}$. 
$i$ denotes $i$-th point and $j$ denotes $j$-th mask features of $k$ original instance mask features.

For the offset branch, we will regress the relative position of each foreground point relative to the centroid of the instance to which it belongs, and we use the same loss function as SoftGroup~\cite{Vu2022SoftGroupF3} for training. The regression loss is as follows:

\begin{equation}
    \label{equ:loss offset}
\mathcal{L}_{\text{offset}} = \frac{1}{\sum_{i=1}^{N} \mathbb{1}_{ \{\bm{s}^i\} } } 
\sum_{i=1}^{N} \mathbb{1}_{ \{\bm{s}^i \} } {||(\bm{o}^{i} - \bm{\hat{o}}^{i})||}_1
\end{equation}

where $\mathbb{1}_{ \{\bm{s}^i\} }$ indicates whether the point belongs to the foreground, $\bm{o}^i$ denotes the predicted centroid offset and $\bm{\hat{o}}^i$ denotes the ground truth centroid offset. We will ignore the points that do not belong to the foreground during training. We can also easily get the centroid position of the instance by adding the centroid offset to the coordinates of the point.

In order to enhance the perception of the instance centroid position of the point-wise feature, we use positional encoding\cite{Fourier} to fuse the centroid features with the original point-wise features

\begin{equation}
    \label{equ: point-wise feature}
f^i = f_{\text{origin}}^i + \text{PE}(p^i + o^i)
\end{equation}

For the mask branch, the branch will generate $k$ initial instance masks $\bm{m}^j \in \mathbb{R}^M$. Each initial instance masks $\bm{m}_{\text{origin}}^j$ can be used to represent the probability that the candidate instance contains each point.

The semantic branch will be used as an auxiliary branch to improve U-Net's perception of semantic information.
The point-wise semantic prediction $\bm{S}$ 
will use the cross entropy loss $\mathcal{L}_{sem} = CE(\bm{S}, \hat{\bm{S}})$ for training. 

\begin{figure*}[t]
    \centering
    \includegraphics[width=0.9\linewidth]{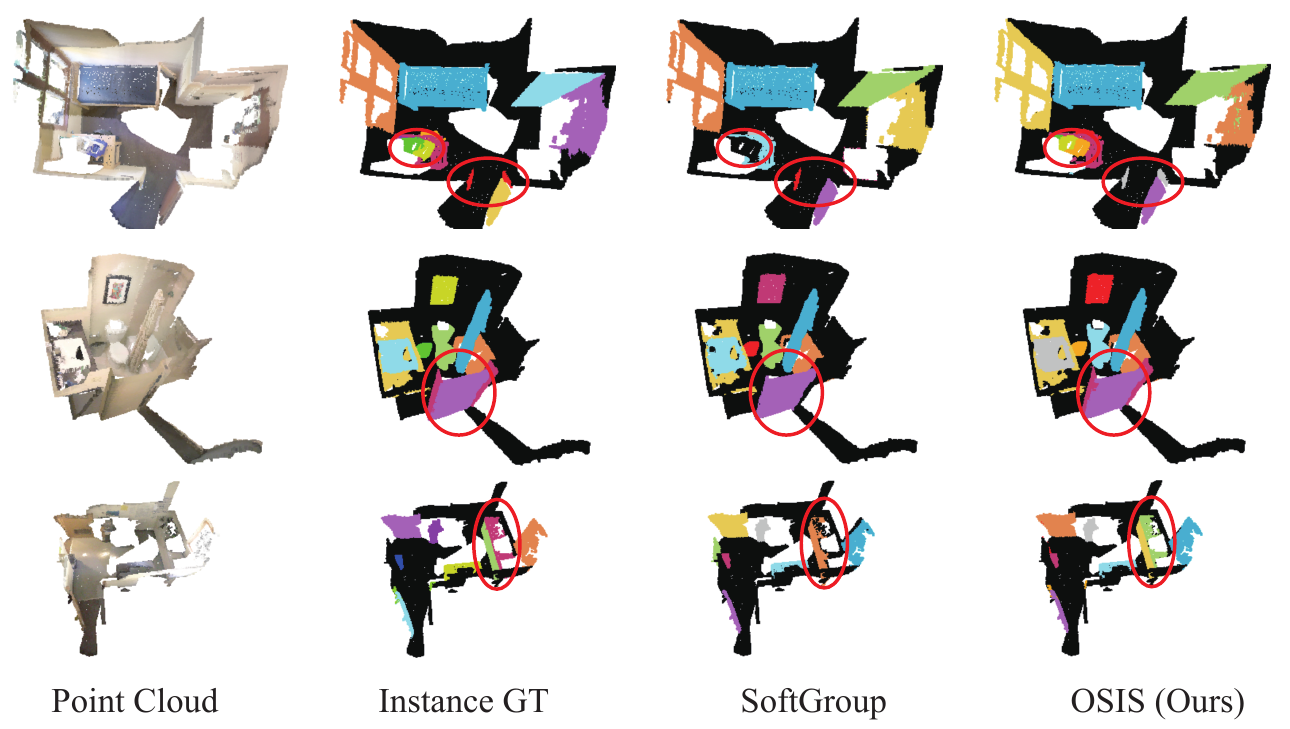}
    \caption{Qualitative comparison results with SoftGroup~\cite{Vu2022SoftGroupF3}. The cases circled with red color show that our network can achieve less error on the instances composed by non-connected parts such as curtains and long desks.}
    \label{fig:qualitative}
\end{figure*}

\subsection{Instance Decoder}

Our instance decoder module mainly consists of two parts: mask-to-feature and feature-to-instance. We will take the point-wise feature $\bm{f}^i$ and initial instance mask $\bm{m}_{\text{origin}}^j$ obtained from the point-wise feature extraction module as the input of the module. 
The module which is implemented through weighted sum and 1x1 dynamic convolution will output the predicted instance masks $\bm{m}^j$ and the semantic categories of the instance $\bm{s}^j$.

\subsubsection{Mask-to-Feature}

In order to obtain instance features and use them as kernel parameters for instance dynamic convolution, we extract instance features through the mask-to-feature module. We first take the instance mask feature obtained in the network as the initial instance mask by adding a sigmoid activation and then apply the initial instance mask to extract the instance feature. The process is as follows:

\begin{gather}
    \bm{f}_{ins}^j = \frac{\sum_{i=1}^{N} {\hat{\bm{m}^j}}^T \cdot \bm{f}^i}{N} 
\\
\hat{\boldsymbol{m}^j} =(1-[\sigma(\boldsymbol{m}^j_{origin}) < \tau])\cdot \sigma(\boldsymbol{m}^j_{origin})
\end{gather}

where $\sigma$ means sigmoid activation for the initial instance mask $\bm{m}_{\text{origin}}^j$, $[\sigma(\boldsymbol{m}^j_{origin}) < \tau]$ is Iverson bracket, $\hat{\bm{m}^j}$ is the instance mask feature after sigmoid activation and thresholding, and $\tau$ is the threshold value. 
By suppressing the contribution of low-confidence points to the instance feature weighting process, the instance feature will focus on the instance foreground area.
$\bm{f}^i$ denotes the point-wise feature. The instance feature $\bm{f}_{ins}^j$ obtained in this way can be used as the kernel parameters for instance dynamic convolution. 

\subsubsection{Feature-to-Instance}

We use 1x1 dynamic convolution to extract the instance mask, which uses the instance feature $\bm{f}_{ins}^j$ as the convolution parameter to perform 1x1 dynamic convolution, and the output instance mask $\bm{m}^j$ will highlight the points belonging to the instance.

In addition, the instance features are leveraged to predict the semantic category of each instance, and a linear layer is employed as the instance semantic classifier. However, in the absence of information exchange between instance features, multiple candidate instances may correspond to the same object, resulting in erroneous predictions. To address this issue, we introduce a background class in the instance classification to classify duplicate instances as background. To enable information interaction among instance features before classification, we incorporate a max pooling operation to extract global features and fuse them into instance features. This simple feature interaction mechanism facilitates accurate identification of duplicate candidates by the instance semantic classifiers

\begin{equation}
    \label{equ:instance feature aligned}
\bm{f}_{pooled}^j = \text{Maxpool}(\bm{f}_{ins}^j) + \bm{f}_{ins}^j
\end{equation}

After the instance feature interaction, we use the linear layer to predict the semantic category $\{\hat{s}^j\}_{j=1}^{k} \in \mathbb{R}^{k \times c}$ of the instance.

\subsection{Bipartite Matching Training}

We use the set prediction method to train the network. First, we define the similarity between the candidate instances obtained by the network and the ground truth instances as the following:

\begin{equation}
    \label{equ:sim}
Q(i, j) = \hat{s}^{i, j} + \alpha \text{Dice}(\bm{m}^i, \bm{m}_{gt}^j)
\end{equation}

where $\hat{s}^{i, j}$ denotes the semantic score that the $i$-th candidate instance belongs to the $j$-th ground truth instance, $\text{Dice}$ denotes the dice coefficient between the predicted instance mask $\bm{m}^i$ and the ground truth instance mask $\bm{m}_{gt}^j$, and $\alpha$ is a hyperparameter. The dice coefficient is defined as follows:

\begin{equation}
    \label{equ:dice}
\text{Dice}(\bm{m}^i, \bm{m}_{gt}^j) = \frac{2 {\bm{m}^i}^T \cdot \bm{m}_{gt}^j}{ {\bm{m}^i}^T \cdot {\bm{m}^i} + {\bm{m}_{gt}^j}^T \cdot \bm{m}_{gt}^j}
\end{equation}

After obtaining the similarity between the candidate instance and the ground truth instance, we regard the assignment problem of the candidate instance and the ground truth as a bipartite graph matching, and solve it with the Hungarian algorithm. We can get several sets of matching pairs $\langle \text{ins}_i, \hat{\text{ins}}_{\pi(i)} \rangle $, and the obtained assignment pairs will guide the training of the network. At the same time, for candidate instances that do not match any target, we will regard them as background classes for training.

The total loss consists of mask loss, semantic classification loss, and point-wise loss.
The point-wise loss $\mathcal{L}_{\text{point}}$ is meanly used to guide the training of the point-wise feature extraction module, which includes the centroid offset regression loss and semantic segmentation loss. 
For semantic classification loss $\mathcal{L}_{\text{cls}}$, we adopt cross-entropy loss for training.
The mask loss $\mathcal{L}_{\text{mask}}$ is shown in Eq~\ref{equ:sem loss}, we use dice loss and focal loss for joint training. The addition of focal loss helps to accelerate the training processing. At the same time, we also use dice loss to improve the accuracy of mask segmentation

\begin{equation}
    \label{equ:sem loss}
\mathcal{L}_{\text{mask}} = \mathcal{L}_{\text{dice}} + \mathcal{L}_{\text{focal}}
\end{equation}

Finally, the total loss is as follows:

\begin{equation}
    \label{equ:total loss}
\mathcal{L} = \mathcal{L}_{\text{mask}} + \mathcal{L}_{\text{cls}} + \mathcal{L}_{\text{point}}
\end{equation}

\subsection{Inference}

At inference processing of our end-to-end 3D instance segmentation network, for a given 3D point cloud, we first voxelize the point cloud and then input it into our network to obtain candidate instances. The predicted instance mask $\bm{m}^i$ represents the confidence score of each point belonging to the $i$-th instance. Simultaneously, we remove the background class and low-confidence instances according to the predicted score of the semantic class, and the whole pipeline does not need any complex post-processing process such as NMS, which breaks through the bottleneck of inference speed improvement.

%% file: sections/4_exp.tex
\section{Experiments}

\noindent \textbf{Datasets and Evaluation Metrics.} 
Our experiments are conducted on the ScanNet dataset~\cite{Dai2017ScanNetR3} which is a richly-annotated dataset of 3D reconstructed indoor scenes. The dataset includes 1613 scans. We use the official train/validate/test split of the dataset, which contains 1200 scans for training, 312 scans for validation, and 100 scans for testing. The evaluation metrics we will report include AP and AP$_{50}$.

\noindent \textbf{Implementation Details.}
Our implementation is based on PyTorch and spconv library~\cite{spconv2022} which is an implement of 3D sparse convolution. We choose Adam optimizer and train for 1000 epochs, the batch size is set to 4, the learning rate is initialized to 0.001, and apply a cosine annealing after 300 epochs. The voxel size is set to 0.02m. During training, we adopt the same data augmentation methods as SoftGroup~\cite{Vu2022SoftGroupF3}.

\subsection{Runtime analysis}
We compared inference time on the ScanNet dataset, Table~\ref{tab:runtime} shows the comparison results of the average inference time. 
We measured the inference speed of our network on the ScanNet v2 validation set using the same GPU with previous works, and reported the average inference time per scene.
Our method is faster than previous methods, and the speedup is 150\% on the validation set compared with SoftGroup~\cite{Vu2022SoftGroupF3}. 
The advantage of our network inference does not come from the optimization of the network structure, but from our simple one-stage approach, which removes the dependence on complex region-growing based clustering, NMS, supervoxel, and other methods in previous 3D instance segmentation networks.

\subsection{Comparison results}
We compare the performance of instance segmentation results on the test set and validation set respectively, Table~\ref{tab:result} and Table~\ref{tab:scannet_test} show the results. Our proposed one-stage network outperforms some previous multi-stage 3D instance segmentation models, compared with 3D-MPA~\cite{Engelmann20203DMPAMA}, our network achieves 4.5 and 3.7 improvement on AP in validation and test set respectively. 
Table~\ref{tab:scannet_test} shows the instance segmentation results of each category in the test set. Compared with the above methods, our network achieves the best results in most classes.

\input{table/time}

\input{table/test_result}

\subsection{Qualitative analysis}
 We qualitatively compare SoftGroup~\cite{Vu2022SoftGroupF3} and our network by visualizing results. As shown in Fig~\ref{fig:qualitative}, our network can well avoid under-segmentation. Meanwhile, for some cases which incorporate the instances consisting of several non-connected parts, such as door frame, SoftGroup tends to predict over-segmented results, while our network can easily make finer predictions.

\subsection{Ablation Study}

Our network employs bipartite matching during training which establishes a one-to-one correspondence between the predicted instance and the ground truth. Thus, our proposed method does not need to add any post-processing methods such as NMS to remove duplicate candidate instances during inference. To prove the above statement, We compare whether to add NMS to the impact of the network results. The results are shown in Table~\ref{tab:nms}, it should be noted that we use matrix NMS instead of the original NMS to achieve faster inference speed. We can find from the results that adding NMS has little effect on the results.

\input{table/val_result.tex}

\input{table/nms.tex}

%% file: table/time.tex
\begin{table}[]
    \small
    \centering
    \setlength{\tabcolsep}{4pt}
    \caption{Comparison on inference time. We measured the inference speed of our network on the ScanNet v2 validation set using the same GPU, and reported the average inference time per scene.}
    \begin{tabular}{llc@{}} \toprule
        Method     & Component time (ms)                    & Total   (ms) \\ \midrule
        & Backbone   (GPU): 2080                   &              \\
        SGPN \cite{Wang2017SGPNSG}      & Group   merging (CPU): 149000            & 158439      \\
        & Block   merging (CPU): 7119              &   \emph{(1148×)}           \\  \midrule
        & Backbone   (GPU): 2083                   &              \\
        ASIS \cite{Wang2019AssociativelySI}      & Mean   shift (CPU): 172711               & 181913       \\
        & Block   merging (CPU): 7119              &     \emph{(1318×)}         \\  \midrule
        & Backbone   (GPU): 1612                   &              \\
        GSPN \cite{Yi2019GSPNGS}      & Point   sampling (GPU): 9559             & 12702        \\
        & Neighbour   search (CPU): 1500           & \emph{(92×)} \\  \midrule
        & Backbone   (GPU): 2083                   &              \\
        3D-BoNet \cite{Yang2019LearningOB}   & SCN   (GPU): 667                         & 9202         \\
        & Block   merging (CPU): 7119              & \emph{(66.68×)} \\  \midrule
        & Backbone   (GPU): 1497                   &              \\
        GICN \cite{Liu2020LearningGI}      & SCN   (GPU): 667                         & 8615         \\
        & Block   merging(CPU): 7119               & \emph{(62.43×)} \\  \midrule
        & Backbone   GPU): 189                     &              \\
        OccuSeg \cite{Han2020OccuSegO3}   & Supervoxel   (CPU): 1202                 & 1904         \\
        & Clustering   (GPU+CPU): 513              &      (13.80×)        \\  \midrule
        & Backbone   (GPU): 128                    &              \\
        PointGroup \cite{Jiang2020PointGroupDP} & Clustering   (GPU+CPU):221               & 452          \\
        & ScoreNet   (GPU): 103                    &      \emph{(3.28×)}        \\  \midrule
        & Backbone (GPU) 125                       &              \\
        SSTNet \cite{liang2021instance}    & Tree network (GPU+CPU): 229              & 428          \\
        & ScoreNet (GPU): 74                       &    \emph{(3.10×)}          \\  \midrule
        & Pointwise   prediction (GPU): 154       &              \\
        HAIS \cite{Chen2021HierarchicalAF}      & Hier. aggr.  (GPU+CPU): 118   & 339          \\
        & Intra-inst. prediction (GPU): 67      &    \emph{(2.46×)}          \\  \midrule
        & Pointwise   prediction (GPU): 152       &              \\
        SoftGroup \cite{Vu2022SoftGroupF3} & Soft grouping (GPU+CPU): 123             & 345          \\
        & Top-down refinement (GPU): 70                &   \emph{(2.50×)}    \\ \midrule
        \textbf{OSIS (Ours)} 
        & Network Prediction (GPU): 138             & \textbf{138}          \\
        \bottomrule
    \end{tabular}
    
    \label{tab:runtime}
\end{table}

%% file: table/test_result.tex
\begin{table*}[]
    \small
    \centering
    \setlength{\tabcolsep}{2.8pt}
    \caption{3D instance segmentation results on ScanNet v2 test set in terms of AP scores on 18 categories. Compared with the above methods, our network achieves the best results in most classes.}
    \begin{tabular}{@{}l|c|cccccccccccccccccc@{}} \toprule
        Method         & AP & \rotatebox[origin=c]{90}{bathtub} & \rotatebox[origin=c]{90}{bed } & \rotatebox[origin=c]{90}{bookshe.} & \rotatebox[origin=c]{90}{cabinet} & \rotatebox[origin=c]{90}{chair} & \rotatebox[origin=c]{90}{counter} & \rotatebox[origin=c]{90}{curtain} & \rotatebox[origin=c]{90}{desk} & \rotatebox[origin=c]{90}{door} & \rotatebox[origin=c]{90}{other} & \rotatebox[origin=c]{90}{picture} & \rotatebox[origin=c]{90}{fridge} & \rotatebox[origin=c]{90}{s. curtain} & \rotatebox[origin=c]{90}{sink} & \rotatebox[origin=c]{90}{sofa} & \rotatebox[origin=c]{90}{table} & \rotatebox[origin=c]{90}{toilet} & \rotatebox[origin=c]{90}{window} \\ \midrule
        3D-SIS \cite{Hou20193DSIS3S}        & 16.1          & 40.7 & 15.5          & 6.8          & 4.3          & 34.6          & 0.1           & 13.4          & 0.5           & 8.8          & 10.6          & 3.7           & 13.5          & 32.1           & 2.8          & 33.9          & 11.6          & 46.6           & 9.3          \\
        3D-BoNet \cite{Yang2019LearningOB}      & 25.3  & 51.9  & 32.4  & 25.1  & 13.7  & 34.5  & 3.1   & 41.9  & 6.9   & 16.2  & 13.1  & 5.2   & 20.2  & 33.8 & 14.7   & 30.1  & 30.3  & 65.1  & 17.8 \\
        MTML \cite{Lahoud20193MTML}          & 28.2  & 57.7  & 38.0  & 18.2  & 10.7  & 43.0  & 0.1   & 42.2  & 5.7   & 17.9  & 16.2  & 7.0   & 22.9  & 51.1 & 16.1   & 49.1  & 31.3  & 65.0  & 16.2          \\
        3D-MPA \cite{Engelmann20203DMPAMA}        & 35.5  & 45.7  & 48.4  & 29.9  & 27.7  & 59.1  & 4.7   & 33.2  & 21.2  & 21.7  & 27.8  & 19.3  & 41.3  & 41.0 & 19.5   & 57.4  & 35.2  & 84.9  & 21.3          \\
        Dyco3D \cite{He2021DyCo3DRI}        & 39.5  & 64.2  & 51.8  & \textbf{44.7}  & 25.9  & \textbf{66.6}  & 5.0   & 25.1  & 16.6  & 23.1  & \textbf{36.2}  & 23.2  & 33.1  & 53.5 & 22.9   & \textbf{58.7}  & \textbf{43.8}  & 85.0  & \textbf{31.7}         \\
        PE \cite{Zhang2019PE}            & 39.6  & 66.7  & 46.7  & 44.6  & 24.3  & 62.4  & 2.2   & \textbf{57.7}  & 10.6  & 21.9  & 34.0  & \textbf{23.9}  & \textbf{48.7}  & 47.5  & 22.5  & 54.1  & 35.0  & 81.8  & 27.3         \\
        PointGroup \cite{Jiang2020PointGroupDP}    & \textbf{40.7}  & 63.9  & 49.6  & 41.5  & 24.3  & 64.5  & 2.1   & 57.0  & 11.4  & 21.1  & 35.9  & 21.7  & 42.8  & \textbf{66.0}  & 25.6  & 56.2  & 34.1  & 86.0  & 29.1       \\
        \textbf{OSIS} (Ours)     & 39.2  & \textbf{77.8}  & \textbf{53.0}  & 22.0  & \textbf{27.8}  & 56.7  & \textbf{8.3}   & 33.0  & \textbf{29.9}  & \textbf{27.0}  & 31.0  & 14.3  & 26.0  & 62.4  & \textbf{27.7}  & 56.8  & 36.1  & \textbf{86.5}  & 30.1 \\ 
        \bottomrule
    \end{tabular}
    
    \label{tab:scannet_test}
\end{table*}

%% file: table/val_result.tex
% include tabularx

\begin{table}[t]
    \centering
    \vspace{5pt}
    \caption{Instance segmentation results on ScanNet v2 validation set and test set.}
    \small
    \begin{tabular}{lccccc} 
        \toprule
        & \multicolumn{2}{c}{val} && \multicolumn{2}{c}{test} \\ 
        \cmidrule{2-3} \cmidrule{5-6}
        Method                                   & AP  & AP$_{50}$  && AP & AP$_{50}$ \\ 
        \midrule
        3D-SIS \cite{Hou20193DSIS3S}             & 18.7   & 35.7       && 16.1   & 38.2         \\
        GSPN \cite{Yi2019GSPNGS}                 & 19.3   & 37.8       && 15.8   & 30.6         \\
        3D-MPA \cite{Engelmann20203DMPAMA}       & 35.5   & 51.9       && 35.5   & 61.1         \\
        PointGroup \cite{Jiang2020PointGroupDP}  & 34.8   & 51.7       && 40.7   & 63.6         \\
        DyCo3D \cite{He2021DyCo3DRI}             & 35.4   & 57.6       && 39.5   & 64.1         \\
        HAIS \cite{Chen2021HierarchicalAF}       & 43.5   & 64.4       && 45.7   & 69.9         \\
        SSTNet \cite{liang2021instance}          & \textbf{49.4}   & 64.3       && \textbf{50.6}   & 69.8         \\
        SoftGroup \cite{Vu2022SoftGroupF3}       & 46.0   & \textbf{67.6}       && 50.4   & \textbf{76.1}         \\
        \midrule
        \textbf{OSIS} (ours)                           & 40.0   & 59.3       && 39.2   & 60.5         \\ 
        \bottomrule 
    \end{tabular}
    
    \label{tab:result}
\end{table}

%% file: table/nms.tex
\begin{table}[]
\centering
\caption{Ablation study of NMS post-processing. The result shows that our network is insensitive to post-processing of removing duplicate instance predictions.}
\label{tab:nms}
\begin{tabular}{|l|c|c|c|}
\hline
Method & AP & AP$_{50}$ & AP$_{25}$ \\ 
\hline
w NMS & 40.3 & 59.4 & 69.7 \\ 
\hline
w/o NMS & 40.0 & 59.3 & 70.0 \\ 
\hline
% \bottomrule
\end{tabular}
\end{table}

%% file: sections/5_conclusion.tex
\section{Conclusion}

In this paper, we introduce a novel one-stage 3D instance segmentation network named OSIS that predicts instance masks directly from the network. In contrast to the conventional multi-stage 3D instance segmentation approaches, our method not only achieves considerable improvement in inference speed but also offers a superior solution to cases where instances comprise multiple disconnected parts, a limitation that was previously challenging to overcome. Furthermore, our network possesses a straightforward and versatile architecture, capable of completing a single frame scene inference within a mere 138ms, making it a potential baseline for real-time 3D instance segmentation.

\section{ACKNOWLEDGMENTS}

This work is supported by the Young Scientists Fund of the National Natural Science Foundation of China (Grant No.62206106). 

%% file: main.bbl
\begin{thebibliography}{10}
\providecommand{\url}[1]{#1}
\csname url@samestyle\endcsname
\providecommand{\newblock}{\relax}
\providecommand{\bibinfo}[2]{#2}
\providecommand{\BIBentrySTDinterwordspacing}{\spaceskip=0pt\relax}
\providecommand{\BIBentryALTinterwordstretchfactor}{4}
\providecommand{\BIBentryALTinterwordspacing}{\spaceskip=\fontdimen2\font plus
\BIBentryALTinterwordstretchfactor\fontdimen3\font minus
  \fontdimen4\font\relax}
\providecommand{\BIBforeignlanguage}[2]{{%
\expandafter\ifx\csname l@#1\endcsname\relax
\typeout{** WARNING: IEEEtran.bst: No hyphenation pattern has been}%
\typeout{** loaded for the language `#1'. Using the pattern for}%
\typeout{** the default language instead.}%
\else
\language=\csname l@#1\endcsname
\fi
#2}}
\providecommand{\BIBdecl}{\relax}
\BIBdecl

\bibitem{Wang2021FCOS3DFC}
T.~Wang, X.~Zhu, J.~Pang, and D.~Lin, ``Fcos3d: Fully convolutional one-stage
  monocular 3d object detection,'' \emph{2021 IEEE/CVF International Conference
  on Computer Vision Workshops (ICCVW)}, pp. 913--922, 2021.

\bibitem{Zhang2020H3DNet3O}
Z.~Zhang, B.~Sun, H.~Yang, and Q.-X. Huang, ``H3dnet: 3d object detection using
  hybrid geometric primitives,'' in \emph{European Conference on Computer
  Vision}, 2020.

\bibitem{Han2020OccuSegO3}
L.~Han, T.~Zheng, L.~Xu, and L.~Fang, ``Occuseg: Occupancy-aware 3d instance
  segmentation,'' \emph{2020 IEEE/CVF Conference on Computer Vision and Pattern
  Recognition (CVPR)}, pp. 2937--2946, 2020.

\bibitem{Chen2021HierarchicalAF}
S.~Chen, J.~Fang, Q.~Zhang, W.~Liu, and X.~Wang, ``Hierarchical aggregation for
  3d instance segmentation,'' \emph{2021 IEEE/CVF International Conference on
  Computer Vision (ICCV)}, pp. 15\,447--15\,456, 2021.

\bibitem{He2021DyCo3DRI}
T.~He, C.~Shen, and A.~van~den Hengel, ``Dyco3d: Robust instance segmentation
  of 3d point clouds through dynamic convolution,'' \emph{2021 IEEE/CVF
  Conference on Computer Vision and Pattern Recognition (CVPR)}, pp. 354--363,
  2021.

\bibitem{Vu2022SoftGroupF3}
T.~Vu, K.~Kim, T.~M. Luu, X.~T. Nguyen, and C.-D. Yoo, ``Softgroup for 3d
  instance segmentation on point clouds,'' \emph{2022 IEEE/CVF Conference on
  Computer Vision and Pattern Recognition (CVPR)}, pp. 2698--2707, 2022.

\bibitem{Jiang2020PointGroupDP}
L.~Jiang, H.~Zhao, S.~Shi, S.~Liu, C.-W. Fu, and J.~Jia, ``Pointgroup: Dual-set
  point grouping for 3d instance segmentation,'' \emph{2020 IEEE/CVF Conference
  on Computer Vision and Pattern Recognition (CVPR)}, pp. 4866--4875, 2020.

\bibitem{Engelmann20203DMPAMA}
F.~Engelmann, M.~Bokeloh, A.~Fathi, B.~Leibe, and M.~Nie{\ss}ner, ``3d-mpa:
  Multi-proposal aggregation for 3d semantic instance segmentation,''
  \emph{2020 IEEE/CVF Conference on Computer Vision and Pattern Recognition
  (CVPR)}, pp. 9028--9037, 2020.

\bibitem{Wang2019AssociativelySI}
X.~Wang, S.~Liu, X.~Shen, C.~Shen, and J.~Jia, ``Associatively segmenting
  instances and semantics in point clouds,'' \emph{2019 IEEE/CVF Conference on
  Computer Vision and Pattern Recognition (CVPR)}, pp. 4091--4100, 2019.

\bibitem{Pang2021SimpleTrackUA}
Z.~Pang, Z.~Li, and N.~Wang, ``Simpletrack: Understanding and rethinking 3d
  multi-object tracking,'' \emph{ArXiv}, vol. abs/2111.09621, 2021.

\bibitem{text2shape}
K.~Chen, C.~B. Choy, M.~Savva, A.~X. Chang, T.~A. Funkhouser, and S.~Savarese,
  ``Text2shape: Generating shapes from natural language by learning joint
  embeddings,'' in \emph{Computer Vision - {ACCV} 2018 - 14th Asian Conference
  on Computer Vision, Perth, Australia, December 2-6, 2018, Revised Selected
  Papers, Part {III}}, 2018, pp. 100--116.

\bibitem{Tang2021Part2WordLJ}
C.~Tang, X.~Yang, B.~Wu, Z.~Han, and Y.~Chang, ``Part2word: Learning joint
  embedding of point clouds and text by matching parts to words,''
  \emph{ArXiv}, vol. abs/2107.01872, 2021.

\bibitem{Liu20223DQueryISAQ}
J.~Liu, T.~He, H.~Yang, R.~Su, J.~Tian, J.~Wu, H.~Guo, K.~Xu, and W.~Ouyang,
  ``3d-queryis: A query-based framework for 3d instance segmentation,''
  \emph{arXiv preprint arXiv:2211.09375}, 2022.

\bibitem{Schult2022Mask3DF3}
J.~Schult, F.~Engelmann, A.~Hermans, O.~Litany, S.~Tang, and B.~Leibe, ``Mask3d
  for 3d semantic instance segmentation,'' \emph{ArXiv}, vol. abs/2210.03105,
  2022.

\bibitem{PointInst}
T.~He, C.~Shen, and A.~van~den Hengel, ``Pointinst3d: Segmenting 3d instances
  by points,'' \emph{ArXiv}, vol. abs/2204.11402, 2022.

\bibitem{Hou20193DSIS3S}
J.~Hou, A.~Dai, and M.~Nie{\ss}ner, ``3d-sis: 3d semantic instance segmentation
  of rgb-d scans,'' \emph{2019 IEEE/CVF Conference on Computer Vision and
  Pattern Recognition (CVPR)}, pp. 4416--4425, 2019.

\bibitem{Yi2019GSPNGS}
L.~Yi, W.~Zhao, H.~Wang, M.~Sung, and L.~J. Guibas, ``Gspn: Generative shape
  proposal network for 3d instance segmentation in point cloud,'' \emph{2019
  IEEE/CVF Conference on Computer Vision and Pattern Recognition (CVPR)}, pp.
  3942--3951, 2019.

\bibitem{Yang2019LearningOB}
B.~Yang, J.~Wang, R.~Clark, Q.~Hu, S.~Wang, A.~Markham, and A.~Trigoni,
  ``Learning object bounding boxes for 3d instance segmentation on point
  clouds,'' \emph{ArXiv}, vol. abs/1906.01140, 2019.

\bibitem{Wu2020SSTNetDM}
X.~Wu, Z.~Xie, Y.~Gao, and Y.~Xiao, ``Sstnet: Detecting manipulated faces
  through spatial, steganalysis and temporal features,'' \emph{ICASSP 2020 -
  2020 IEEE International Conference on Acoustics, Speech and Signal Processing
  (ICASSP)}, pp. 2952--2956, 2020.

\bibitem{Chen2019DynamicCA}
Y.~Chen, X.~Dai, M.~Liu, D.~Chen, L.~Yuan, and Z.~Liu, ``Dynamic convolution:
  Attention over convolution kernels,'' \emph{2020 IEEE/CVF Conference on
  Computer Vision and Pattern Recognition (CVPR)}, pp. 11\,027--11\,036, 2019.

\bibitem{Yang2019CondConvCP}
B.~Yang, G.~Bender, Q.~V. Le, and J.~Ngiam, ``Condconv: Conditionally
  parameterized convolutions for efficient inference,'' in \emph{Neural
  Information Processing Systems}, 2019.

\bibitem{wang2020solov2}
X.~Wang, R.~Zhang, T.~Kong, L.~Li, and C.~Shen, ``Solov2: Dynamic and fast
  instance segmentation,'' \emph{Advances in Neural information processing
  systems}, vol.~33, pp. 17\,721--17\,732, 2020.

\bibitem{Tian2020ConditionalCF}
Z.~Tian, C.~Shen, and H.~Chen, ``Conditional convolutions for instance
  segmentation,'' in \emph{European Conference on Computer Vision}, 2020.

\bibitem{Cheng2022SparseIA}
T.~Cheng, X.~Wang, S.~Chen, W.~Zhang, Q.~Zhang, C.~Huang, Z.~Zhang, and W.~Liu,
  ``Sparse instance activation for real-time instance segmentation,''
  \emph{2022 IEEE/CVF Conference on Computer Vision and Pattern Recognition
  (CVPR)}, pp. 4423--4432, 2022.

\bibitem{Dong2021SOLQSO}
B.~Dong, F.~Zeng, T.~Wang, X.~Zhang, and Y.~Wei, ``Solq: Segmenting objects by
  learning queries,'' in \emph{Neural Information Processing Systems}, 2021.

\bibitem{Cheng2021MaskFormer}
B.~Cheng, A.~G. Schwing, and A.~Kirillov, ``Per-pixel classification is not all
  you need for semantic segmentation,'' in \emph{Neural Information Processing
  Systems}, 2021.

\bibitem{Cheng2022Mask2Former}
B.~Cheng, I.~Misra, A.~G. Schwing, A.~Kirillov, and R.~Girdhar,
  ``Masked-attention mask transformer for universal image segmentation,''
  \emph{2022 IEEE/CVF Conference on Computer Vision and Pattern Recognition
  (CVPR)}, pp. 1280--1289, 2022.

\bibitem{Zhang2021KNetTU}
W.~Zhang, J.~Pang, K.~Chen, and C.~C. Loy, ``K-net: Towards unified image
  segmentation,'' in \emph{Neural Information Processing Systems}, 2021.

\bibitem{Erhan2013ScalableOD}
D.~Erhan, C.~Szegedy, A.~Toshev, and D.~Anguelov, ``Scalable object detection
  using deep neural networks,'' \emph{2014 IEEE Conference on Computer Vision
  and Pattern Recognition}, pp. 2155--2162, 2013.

\bibitem{Carion2020DETR}
N.~Carion, F.~Massa, G.~Synnaeve, N.~Usunier, A.~Kirillov, and S.~Zagoruyko,
  ``End-to-end object detection with transformers,'' \emph{ArXiv}, vol.
  abs/2005.12872, 2020.

\bibitem{sparseconv}
B.~Graham, M.~Engelcke, and L.~van~der Maaten, ``3d semantic segmentation with
  submanifold sparse convolutional networks,'' \emph{2018 IEEE/CVF Conference
  on Computer Vision and Pattern Recognition}, pp. 9224--9232, 2018.

\bibitem{ronneberger2015unet}
O.~Ronneberger, P.~Fischer, and T.~Brox, ``U-net: Convolutional networks for
  biomedical image segmentation,'' in \emph{International Conference on Medical
  image computing and computer-assisted intervention}.\hskip 1em plus 0.5em
  minus 0.4em\relax Springer, 2015, pp. 234--241.

\bibitem{Fourier}
M.~Tancik, P.~P. Srinivasan, B.~Mildenhall, S.~Fridovich{-}Keil, N.~Raghavan,
  U.~Singhal, R.~Ramamoorthi, J.~T. Barron, and R.~Ng, ``Fourier features let
  networks learn high frequency functions in low dimensional domains,'' in
  \emph{NeurIPS}, 2020.

\bibitem{Dai2017ScanNetR3}
A.~Dai, A.~X. Chang, M.~Savva, M.~Halber, T.~A. Funkhouser, and M.~Nie{\ss}ner,
  ``Scannet: Richly-annotated 3d reconstructions of indoor scenes,'' \emph{2017
  IEEE Conference on Computer Vision and Pattern Recognition (CVPR)}, pp.
  2432--2443, 2017.

\bibitem{spconv2022}
S.~Contributors, ``Spconv: Spatially sparse convolution library,'' 2022.

\bibitem{Wang2017SGPNSG}
W.~Wang, R.~Yu, Q.~Huang, and U.~Neumann, ``Sgpn: Similarity group proposal
  network for 3d point cloud instance segmentation,'' \emph{2018 IEEE/CVF
  Conference on Computer Vision and Pattern Recognition}, pp. 2569--2578, 2017.

\bibitem{Liu2020LearningGI}
S.-H. Liu, S.~Yu, S.-C. Wu, H.-T. Chen, and T.-L. Liu, ``Learning gaussian
  instance segmentation in point clouds,'' \emph{ArXiv}, vol. abs/2007.09860,
  2020.

\bibitem{liang2021instance}
Z.~Liang, Z.~Li, S.~Xu, M.~Tan, and K.~Jia, ``Instance segmentation in 3d
  scenes using semantic superpoint tree networks,'' \emph{2021 IEEE/CVF
  International Conference on Computer Vision (ICCV)}, pp. 2763--2772, 2021.

\bibitem{Lahoud20193MTML}
J.~Lahoud, B.~Ghanem, M.~Pollefeys, and M.~R. Oswald, ``3d instance
  segmentation via multi-task metric learning,'' \emph{2019 IEEE/CVF
  International Conference on Computer Vision (ICCV)}, pp. 9255--9265, 2019.

\bibitem{Zhang2019PE}
B.~Zhang and P.~Wonka, ``Point cloud instance segmentation using probabilistic
  embeddings,'' \emph{2021 IEEE/CVF Conference on Computer Vision and Pattern
  Recognition (CVPR)}, pp. 8879--8888, 2019.

\end{thebibliography}
